\documentclass[conference]{IEEEtran}
\IEEEoverridecommandlockouts
\usepackage{cite}
\usepackage{amsmath,amssymb,amsfonts}
\usepackage{algorithmic}
\usepackage{graphicx}
\usepackage{textcomp}
\usepackage{xcolor}
\def\BibTeX{{\rm B\kern-.05em{\sc i\kern-.025em b}\kern-.08em
    T\kern-.1667em\lower.7ex\hbox{E}\kern-.125emX}}

\usepackage{graphicx}
\usepackage{subcaption}
\usepackage{booktabs}  

\usepackage{wrapfig}

\usepackage{amsfonts}
\usepackage{amssymb}
\usepackage{amsmath}
\usepackage{multirow}
\let\vec\boldsymbol

\usepackage[hidelinks]{hyperref}

\usepackage{multirow}
\usepackage{amssymb}
\usepackage[english]{babel}
\usepackage{hyphenat}
\makeatletter
\newcommand{\ssymbol}[1]{$^{\@fnsymbol{#1}}$}
\makeatother

\let\vec\boldsymbol

\newcommand{\R}{\mathbb{R}}
\begin{document}

\title{Exploring Semantic Relationships for \\ Unpaired Image Captioning}

\author{Fenglin Liu\textsuperscript{1}, Meng Gao\textsuperscript{2}, Tianhao Zhang\textsuperscript{3}, Yuexian Zou\textsuperscript{1}\\
\IEEEauthorblockA{\textsuperscript{1}ADSPLAB, School of ECE, Peking University}
\IEEEauthorblockA{\textsuperscript{2}School of ICE, Beijing University of Posts and Telecommunications}
\IEEEauthorblockA{\textsuperscript{3}Beijing Foreign Studies University}
\IEEEauthorblockA{{\tt \{fenglinliu98, zouyx\}@pku.edu.cn, gaomeng@bupt.edu.cn, wszth666@bfsu.edu.cn}}
}

\maketitle

\begin{abstract}
Recently, image captioning has aroused great interest in both academic and industrial worlds. Most existing systems are built upon large-scale datasets consisting of image-sentence pairs, which, however, are time-consuming to construct. In addition, even for the most advanced image captioning systems, it is still difficult to realize deep image understanding. In this work, we achieve unpaired image captioning by bridging the vision and the language domains with high-level semantic information. The motivation stems from the fact that the semantic concepts with the same modality can be extracted from both images and descriptions. To further improve the quality of captions generated by the model, we propose the \textit{Semantic Relationship Explorer}, which explores the relationships between semantic concepts for better understanding of the image. Extensive experiments on MSCOCO dataset show that we can generate desirable captions without paired datasets. Furthermore, the proposed approach boosts five strong baselines under the paired setting, where the most significant improvement in CIDEr score reaches 8\%, demonstrating that it is effective and generalizes well to a wide range of models\footnote{This paper was originally published at ICDM 2019 \cite{fenglin2019exploring}.}.

\end{abstract}

\begin{IEEEkeywords}
vision and language, unpaired training data, image captioning, deep neural network
\end{IEEEkeywords}

\section{Introduction}
Image captioning has drawn remarkable attention in both natural language processing and computer vision. The task, which combines image understanding and language generation, is tough yet practical. Above all, it has various kinds of applications such as human-robot interaction \cite{das2017visual}, text-based image retrieval \cite{liu2018retrieval} and helping visually impaired people see \cite{wu2017automatic}, among others. The deep neural networks, particularly those based on the encoder-decoder framework \cite{xu2015show,you2016image,yao2017boosting,fenglin2018simnet,anderson2018bottom,fenglin2020federated,yao2018exploring,fenglin2019GLIED,Yang2019SGAE}, have achieved great success in advancing the state-of-the-art.

Despite the impressive achievements of the deep learning frameworks, they unduly rely on large-scale paired data, which is not easy to access in the real world. Especially, when it comes to the Non-English caption systems, image-caption pairs are labor-intensive to obtain. In recent years, unsupervised encoder-decoder models have been proposed for neural machine translation \cite{Artetxe2018UnsupervisedNMT,Lample2018UnsupervisedNMT,Lample2018Phrase-Based}. Typically, the source language and the target language are mapped into a common latent space, where sentences with the same semantic meaning are well aligned, thus the unsupervised translation can be carried out. For the task of image captioning, due to the great disparities between the vision and the language domains, and distinct features of the modalities,  unsupervised learning is even more challenging. Although unpaired image-to-sentence translation has been explored in the literature, the training of neural models on unpaired datasets has not yet been conducted at a semantic level.

\begin{figure}[t]

\centering
\includegraphics[width=1\linewidth]{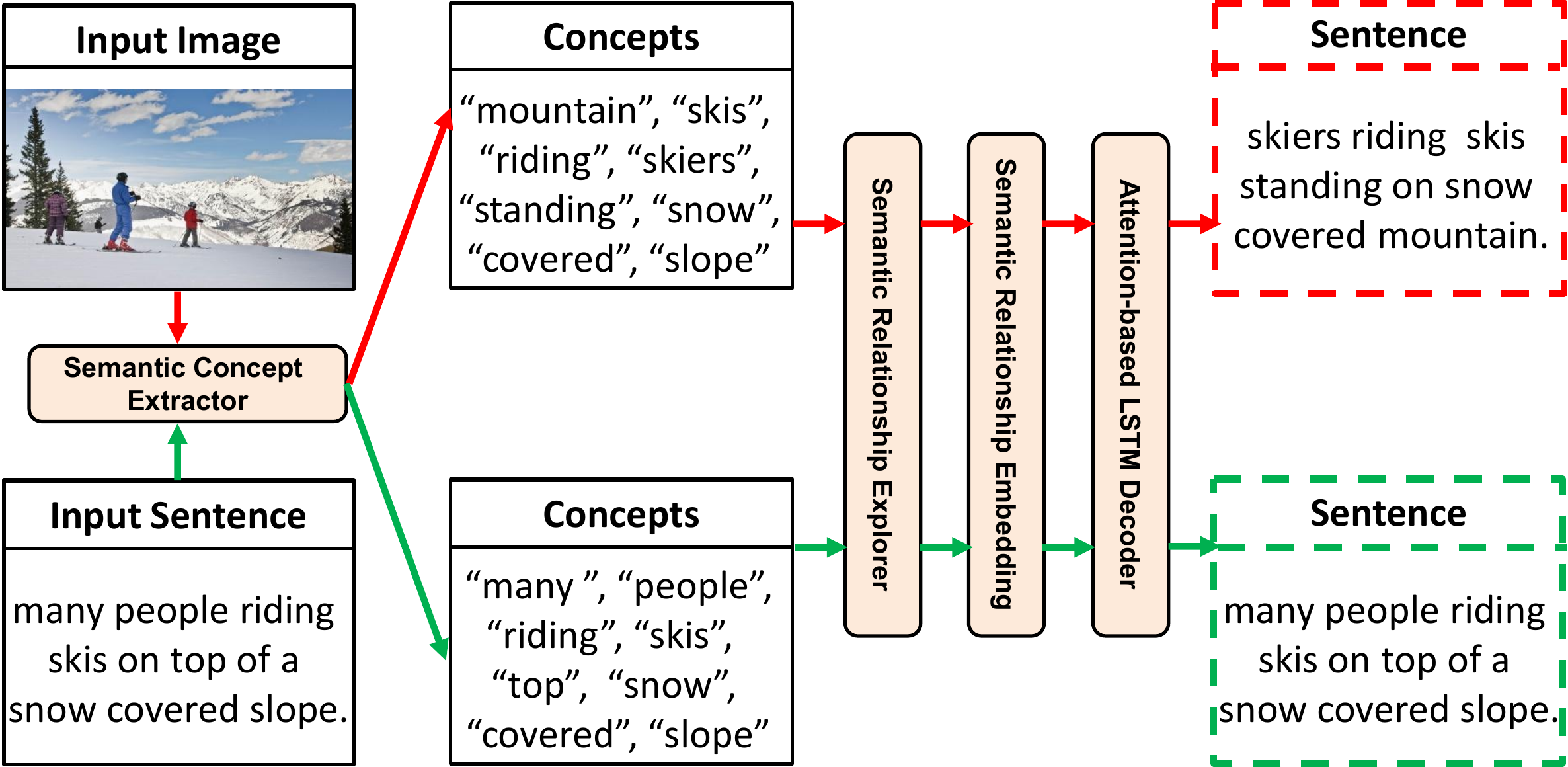}

\caption{Illustration of our proposed framework. Our framework consists of a Semantic Concept Extractor, a Semantic Relationship Explorer, a \textit{Semantic Relationship Embedding} and an Attention-based Sentence Decoder.}
\label{fig:introduction}
\vspace{-10pt}
\end{figure}

Recently, there have been a few works \cite{gu2018unpaired,Feng2018Unsupervised_ic,gu2019unpaired,Laina2019Towards,Gao2020Unsupervised,Cao2020Interactions,Kim2019Scarce} trying to reduce the dependence of image captioning models on paired data.
Specifically, \cite{gu2018unpaired} proposed a method to generate captions in a central language (Chinese) and then translate them into a target language (English), without requirement for image-caption pairs for the target language. However, a paired image captioning corpus for the source language and a paired corpus for translation are still indispensable. 
\cite{Feng2018Unsupervised_ic} connected the visual and the textual modalities with visual objects (e.g., \textit{dog}, \textit{mirror}) and an adversarial manner. It achieved unsupervised image captioning without any image-caption pairs. Nevertheless, their method relied on the recognized visual objects to decide whether the generated captions are image-related, which necessitated complex models and schemes to obtain higher-quality image captions, e.g. image reconstruction \cite{Feng2018Unsupervised_ic}, sentence reconstruction \cite{Vincent2008autoencoders} and adversarial learning \cite{goodfellow2014GAN}.  Moreover, the visual object based approach does not consider attributive  words (e.g. attributes (\textit{small}), relations (\textit{standing}) and color (\textit{white})), which could help improve the image representation.
\cite{gu2019unpaired} used the scene graph to bridge the gap between the visual and the textual modalities. Since the scene graph constructed a series of semantic relationship information, the model achieved good results. However, in order to construct the semantic relationships, they needed to use Faster-RCNN \cite{ren2015faster} as the object detector, MOTIFS \cite{Zellers2018Motifs} as the relationship detector, and an additional classifier for attribute identification \cite{Yang2019SGAE} for generating the image scene graph.
In all, although existing methods for unsupervised image captioning have made appreciable progress, they are hard to implement and still far from real world applications.

In this paper, we bridge the gap between images and captions with semantic concepts, which explicitly represent high-level information \cite{wu2016what}. To solve the challenges posed by different characteristics between visual and textual modalities, we propose an approach to explore the relationships between individual semantic concepts. The proposed framework (Figure~\ref{fig:introduction}) includes a \textit{Semantic Concept Extractor}, a \textit{Semantic Relationship Explorer}, a \textit{\textit{Semantic Relationship Embedding}} and an \textit{Attention-based Sentence Decoder}. To begin with, our model extracts relevant semantic concepts from the images and the captions, respectively, which is the \textit{Semantic Concept Extractor}. Then, the \textit{Semantic Relationship Explorer}, which consists of an \textit{``Attribute'' Aggregator}, an \textit{``Object'' Aggregator} and a \textit{``Relation'' Aggregator}, explores effective semantic relationships among the extracted concepts from three aspects, i.e., ``attribute'', ``object'' and ``relation''. Meanwhile, the \textit{\textit{Semantic Relationship Embedding}} can extract principle features of the semantic relationships from different parts, which strengthens the expressive ability of relationship features. As a result, characteristic information  of the semantic concepts is fully exploited and the representation of the image concept features is getting closer to the textual modality  and, as input to the decoder, adopted effectively to generate meaningful sentences.
Our main contributions are as follows:

\begin{itemize}
    \item We propose a semantic based framework for unpaired image captioning. The framework can effectively and automatically use the semantic information to generate high quality image captions by bridging the gap between the vision and the language domains.
    \item To explore the relationships between individual semantic concepts, we propose a \textit{Semantic Relationship Explorer}, which can extract both coarse-grained and fine-grained semantic relationships to help the model generate more accurate and meaningful captions.
    \item Our approach proves to be effective by the experiments on the MSCOCO image captioning dataset. We outperforms the latest unpaired image captioning systems. More encouragingly, our method also makes improvements of up to 8\% in terms of CIDEr score for paired image captioning model, when equipping it with our approach.
\end{itemize}

\section{Related Work}

\subsection{Image Captioning}
Recently, many neural model systems have been proposed for image captioning \cite{xu2015show,you2016image,yao2017boosting,fenglin2018simnet,anderson2018bottom,fenglin2020federated,yao2018exploring,fenglin2019GLIED,Yang2019SGAE}. The most advanced methods \cite{anderson2018bottom,fenglin2020federated,yao2018exploring,fenglin2019GLIED,Yang2019SGAE} rely on the encoder-decoder framework and combine the attention mechanism \cite{xu2015show} to transform the image into coherent caption. However, the paired image-sentence data for training these models is expensive to collect. The first model under the unsupervised training setting, built by \cite{gu2018unpaired}, aims to take the pivot language to connect source image and target caption. Although image and target language caption pairs are not used, their method relies on image-pivot pairs and a pivot-target parallel translation corpus. \cite{Feng2018Unsupervised_ic} and \cite{gu2019unpaired} align between the vision and the language domain in an adversarial way, so they don't need any paired image-sentence data. Also, using a scarce amount of paired data as a bridge in the form of semi-supervised learning has been proposed by \cite{Kim2019Scarce}. In contrast to the approaches mentioned above, our approach attempts to regard semantic concepts as a bridge and explore the semantic relationships among them, which has not been well studied yet.

\subsection{Unsupervised Machine Translation}
Unsupervised image captioning is essentially similar to unsupervised machine translation. Nevertheless, due to the apparently different characteristics of the image and text modality, unpaired image-to-sentence translation is more challenging than unpaired sentence-to-sentence translation.

\subsection{Exploring Semantic Relationships}
A most recent advancement \cite{yao2018exploring,Yang2019SGAE} attempts to use graph networks to explicitly explore visual relationships by encoding scene graphs, which model the spatial and semantic relationships of image regions based on visual features. However, they built an additional model to explicitly predict the relationships between visual objects with extra annotated data. Instead, our model associates the semantic concepts based on attention, and the relationships are implicitly modelled as weighted combinations and trained with the captioning model.

\begin{figure*}[t]

\centering
\includegraphics[width=0.8\linewidth]{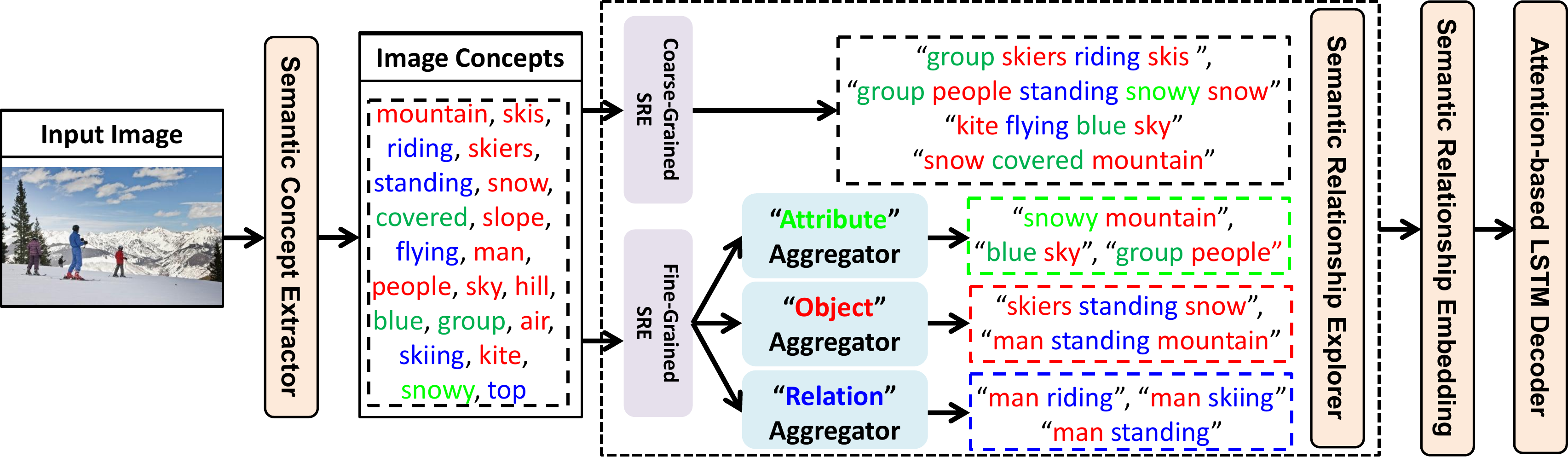}

\caption{Illustration of the proposed framework. The semantic concepts are extracted by an extractor \cite{fang2015captions}. The color \textit{Green} denotes the attribute words, the color \textit{Red} denotes the object words and \textit{Blue} color denotes the relation words. The \textit{Semantic Relationship Explorer} consists of a \textit{Coarse-Grained Semantic Relationship Explorer} and a \textit{Fine-Grained Semantic Relationship Explorer}. The latter one includes three sub-modules: a 
\textit{``Attribute'' Aggregator}, an \textit{``Object'' Aggregator} and a \textit{``Relation'' Aggregator}, then adaptively explores the relationships among the extracted semantic concepts. The proposed approach could achieve deep semantic understanding of the image and in turn generate more complete and coherent captions.}
\label{fig:approach}
\end{figure*}

\section{The Proposed Framework}

As shown in Figure~\ref{fig:approach}, the proposed framework consists of four main modules: (1) \textit{Semantic Concept Extractor}: this module is able to extract the semantic concepts from images or sentences; (2) \textit{Semantic Relationship Explorer} (Figure~\ref{fig:SRE}): since the extracted semantic concepts are independent and not associated with each other, e.g., the three words \textit{riding}, \textit{boy} and \textit{bike}, this module can associate these words together as a phrase \textit{boy riding bike} to represent a complete semantics; (3) \textit{Semantic Relationship Embedding}: this module can strengthen the expressive ability of relationship features by extracting different features of the explored semantic relationships from different parts; (4) Through the above three steps, we are able to provide rich semantic relationship information which is beneficial for the fourth module, i.e., \textit{Attention-based Sentence Decoder}, to generate the complete and coherent image captions under the unpaired setting. In the following parts of this section, we will describe these steps in detail.

\subsection{Semantic Concept Extractor}

In recent studies \cite{fang2015captions,wu2016what,yao2017boosting,Lu2018R-VQA,yu2017Multi-level}, the semantic concepts are introduced to represent the explicit high-level information of the image \cite{wu2016what}. In particular, it consists of a series of words, including attributes (e.g., \textit{young}, \textit{black}), objects (e.g., \textit{woman}, \textit{shirt}) and relationships (e.g., \textit{sitting}, \textit{wearing}). In this case, these semantic concept words for an image can be generated by a \textit{Semantic Concept Extractor} \cite{wu2016what,yao2017boosting}. In implementation, we adopt a weakly-supervised method of Multiple Instance Learning (MIL) \cite{zhang2006multiple} to build the \textit{Semantic Concept Extractor}, following \cite{fang2015captions}.
Technically, for each concept, MIL takes the ``positive'' and ``negative'' bags of bounding boxes as input sets, where every bag corresponds to an image.  If the current concept is in the semantic label (e.g. description and scene graph) of the image, the bag is regarded to be positive, otherwise it is said to be negative.
MIL iteratively selects the positive instances to train an extractor for the semantic concepts. 
Due to limited space, please refer to \cite{fang2015captions} for detailed explanation. 
Other semantic concept extracting approaches can also be used and may produce better results, which, however, are not the main focuses of this work.

To extract the semantic concepts of a sentence, we directly use the words, which appear in both the ground truth and the pre-defined semantic concepts vocab, as semantic concepts.

In addition, because semantic concepts are discrete word tokens, we adapt a word embedding to project them into vectors $\vec{C}_g = \{\vec{c}_1, \vec{c}_2, \ldots, \vec{c}_{N_g}\}, \vec{c}_i \in \R^e$, where $e$ is the dimension of word embedding. Then all words, words containing only ``attribute'', words containing only ``object'' and words containing only ``relation'' are respectively represented as global semantic concepts $\vec{C}_g \in \R^{N_g \times e}$, attribute semantic concepts $\vec{C}_a \in \R^{N_a \times e}$, object semantic concepts $\vec{C}_o \in \R^{N_o \times e}$ and relation semantic concepts $\vec{C}_r \in \R^{N_r \times e}$, where $N_g$, $N_a$, $N_o$ and $N_r$ represent the number of words per type.

\subsection{Semantic Relationship Explorer}
\begin{figure*}[t]

\centering
\includegraphics[width=0.8\linewidth]{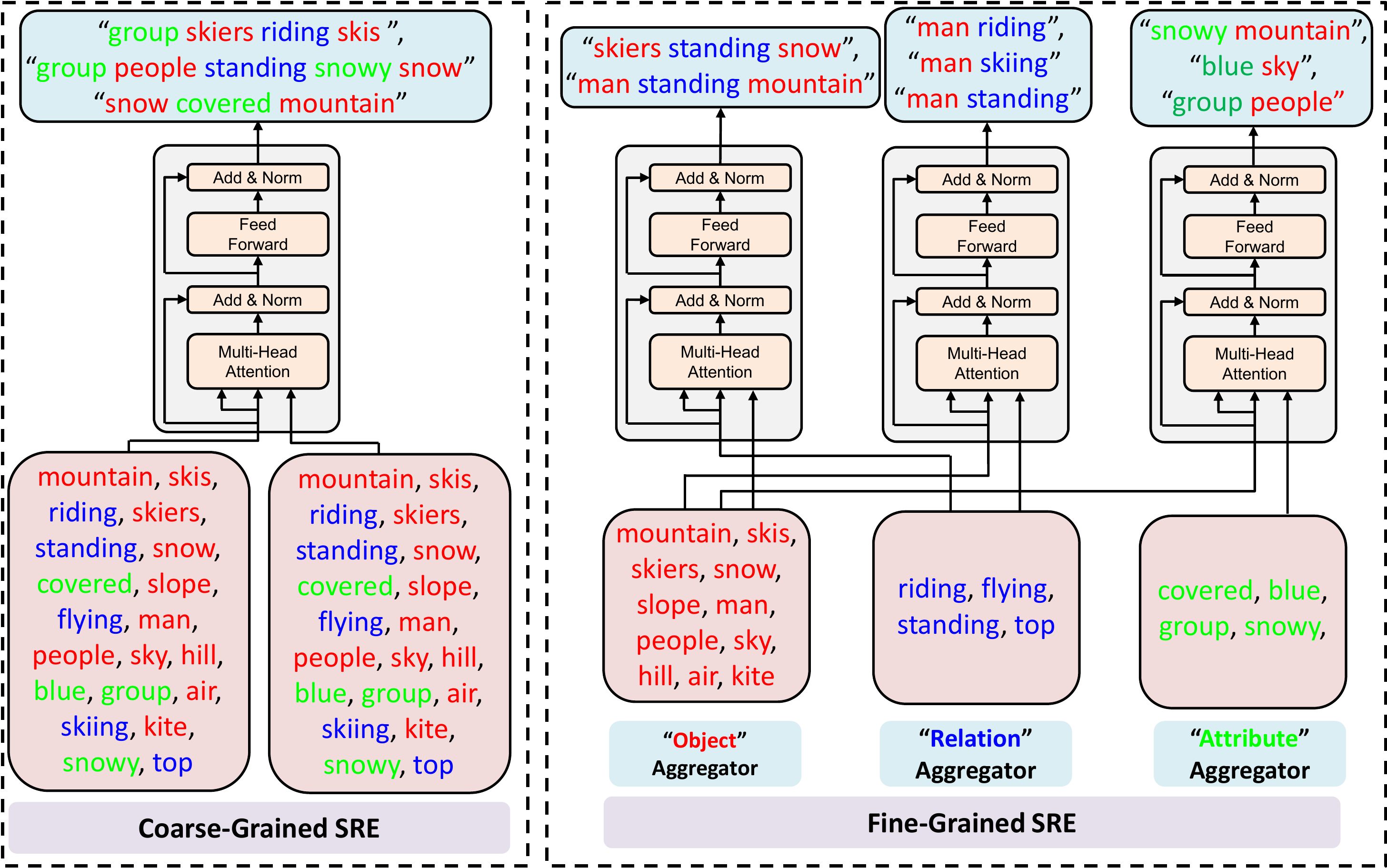}

\caption{Illustration of the proposed Semantic Relationship Explorer. The color \textit{Green} denotes the Attribute words, the color \textit{Red} denotes the Object words and \textit{Blue} color denotes the Relation words.}
\label{fig:SRE}
\end{figure*}

In this section, we will first introduce the method we use to extract the semantic relationships - MultiHead Attention, then we will introduce the proposed \textit{Semantic Relationship Explorer}, Figure~\ref{fig:SRE} shows a sketch.

\subsubsection{MultiHead Attention}

In order to extract the semantic relationships, we adopt multi-head attention \cite{ashish2017attention}, which can calculate the association weights between each pair of the given semantic concepts \cite{fenglin2019mia}. The following multi-head attention consists of $\vec{N}$ parallel heads and each head is represented as a scaled dot-product attention. 
\begin{equation}
\text{Att}(\text{Q,K,V}) = \text{softmax}\left(\frac{\text{Q}\text{W}^\text{Q}(\text{K}\text{W}^\text{K})^{T}}{\sqrt{{d}_{k}}}\right)\text{V}\text{W}^\text{V}
\label{eq.1}
\end{equation}
where $\text{Q} \in \mathbb{R}^{l \times d_h}$, $\text{K} \in \mathbb{R}^{{k} \times d_h}$ and $\text{V} \in \mathbb{R}^{{k} \times d_h}$ represent respectively the query matrix, the key matrix and the value matrix. $\text{W}^\text{Q}, \text{W}^\text{K}, \text{W}^\text{V} \in \mathbb{R}^{d_h \times d_k}$ are learnable parameters and ${d}_{k} = d_h / {n} $, where $n$ represents the number of heads.

Following the multi-head attention is a fully-connected feed-forward network, which is defined as follows:
\begin{equation}
\text{FFN}(x) = \text{ReLU}(x\text{W}_\text{f}+{b}_\text{f})\text{W}_\text{ff}+{b}_\text{ff} 
\end{equation}
where $\text{W}_\text{f}$ and $\text{W}_\text{ff}$ denote matrices for linear transformation; $b_\text{f}$ and ${b}_\text{ff}$ represent the bias terms.

We use the attention-based method to make each semantic concept adaptively find the most relevant attributes, objects and relations, and establish related relationships, thus completing the exploration process of the semantic relationships.

\subsubsection{Coarse-Grained Semantic Relationship Explorer}
The \textit{Coarse-Grained SRE} is a simple and intuitive way to explore the semantic relationships. The relationships are built via multi-head attention on the global semantic concepts $\vec{C}_g$. Particularly, we use $\vec{C}_g$ as Query, Key and Value simultaneously. The result turns out to be a set of attended semantic concepts:
\begin{equation}
\vec{R}_{g} = 
\text{FFN}(\text{MultiHead}(\vec{C}_{g},\vec{C}_{g},\vec{C}_{g}))
\end{equation}
with \textit{Coarse-Grained SRE}, all semantic concepts can attend to other semantic concepts related to them, making it easy to capture global relationships quickly. However, the semantic relationships extracted by \textit{Coarse-Grained SRE} are coarse-grained, because for any kind of semantic concept as a Query, it is not necessary to treat all words as Key and Value, i.e., there is no need to attend to the most relevant among all words. For example, when an ``attribute'' word is used as a Query, it does not need to pay attention to the ``relation'' words. This is because the ``attribute'' words are usually used to describe a specific object, while the ``relation'' words are usually used to describe the relationship among two or more objects, that is, the association between the ``attribute'' words and the ``relation'' words is not very strong. Therefore, the semantic relationships explored by \textit{Coarse-Grained SRE} are not good due to the noise introduced by many unrelated words.

\subsubsection{Fine-Grained Semantic Relationship Explorer}

In order to solve the problems of the \textit{Coarse-Grained SRE} and to extract more fine-grained and more precise semantic relationships, we propose the \textit{Fine-Grained Semantic Relationship Explorer} (\textit{Fine-Grained SRE}). Inspired by the proposed attribute embedding, object embedding and relationship embedding in \cite{Yang2019SGAE}, we propose \textit{``Attribute'' Aggregator}, \textit{``Object'' Aggregator} and \textit{``Relation'' Aggregator}, which use attribute words, object words and relation words, respectively, as Value and Key. In addition, we get finer and more accurate semantic relationships by using the words of the category most relevant to them as Query.

\paragraph{``Attribute'' Aggregator}
A specific object may have multiple attributes. Therefore, we have to focus on the attributes that are most relevant to it from a set of attribute semantic concepts. In this case, according to the attention theorem, the object semantic concepts $\vec{C}_o$ are the Queries, and the attribute semantic concepts $\vec{C}_a$ are the Keys and Values. Consequently, the result turns out to be a set of attended attribute semantic concepts:  
\begin{equation}
\vec{R}_{a} = 
\text{FFN}(\text{MultiHead}(\vec{C}_{o},\vec{C}_{a},\vec{C}_{a}))
\end{equation}
Through this formula, we can get detailed object-attribute relationship information.

\paragraph{``Object'' Aggregator}

For a particular relation concept, we tend to focus on seeking for two objects that are related to each other to form the object-relationship-object, e.g., \textit{women-wearing-shirt}. Therefore, for the \textit{``Object'' Aggregator}, the relation semantic concepts should focus on the object semantic concepts. The relation semantic concepts $\vec{C}_r$ serve as Queries, and the object semantic concepts $\vec{C}_o$ serve as Keys and Values, which can be defined as follows:
\begin{equation}
\vec{R}_{o} = 
\text{FFN}(\text{MultiHead}(\vec{C}_{r},\vec{C}_{o},\vec{C}_{o}))
\end{equation}
Thus, we can get detailed object-relation-object relationship information by the \textit{``Object'' Aggregator}.

\paragraph{``Relation'' Aggregator}

A specific object may have many relations with other objects, so we need to focus on the relation that is most relevant to it from a set of relation semantic concepts, e.g., \textit{woman-eating, woman-holding, woman-wearing} and so on. Therefore, for \textit{``Relation'' Aggregator}, we consider what relationships they are attached to. Accordingly, the object semantic concepts $\vec{C}_o$ should pay attention to the relation semantic concepts $\vec{C}_r$, which is defined as follows:
\begin{equation}
\vec{R}_{r} = 
\text{FFN}(\text{MultiHead}(\vec{C}_{o},\vec{C}_{r},\vec{C}_{r}))
\end{equation}
where we can get object-relation relationship information.

\subsection{\textit{Semantic Relationship Embedding}} \label{sec:embedding}

For the semantic concepts of each image or sentence, through \textit{Coarse-Grained SRE} and \textit{Fine-Grained SRE}, we can get coarse-grained relationships $\vec{R}_g \in \R^{N_g \times e}$ and fine-grained relationships $\vec{R}_a \in \R^{N_a \times e}$, $\vec{R}_o \in \R^{N_o \times e}$ and $\vec{R}_r \in \R^{N_r \times e}$, respectively. These semantic relationships are of different quantity and contain different information. So in order to have the same number of features for each type of semantic relationships and to encourage the diversity of extracted main features for decoding effectively, we use a self-attention mechanism provided by \cite{lin2017Embedding}. This mechanism can be defined as follows:
\begin{align}
\vec{A} &= 
\text{softmax}(\vec{W}_{s_2}\text{tanh}(\vec{W}_{s_1}{\vec{R}_k}^T)) \\
&\vec{M}_k = \vec{A}\vec{R}_k, k \in \{g,a,o,r\}
\end{align}
where $\vec{W}_{s_1} \in \R^{d_a \times e}$ and $\vec{W}_{s_2} \in \R^{r \times d_a}$ are learnable parameters; \textit{r} stands for the number of extracted features in different parts. Through the formula, we can get different semantic relationship embeddings, i.e.,  $\vec{M}_k \in \R^{r \times e} (k \in \{g,a,o,r\})$, which share the same dimensions. For coarse-grained relationships embedding, $\vec{M}_c = \vec{M}_g$. We combine $\vec{M}_a$, $\vec{M}_o$ and $\vec{M}_r$ to obtain the fine-grained relationships embedding, $\vec{M}_f = [\vec{M}_a;\vec{M}_o;\vec{M}_r]$, where [;] stands for concatenation operation.

\subsection{Attention-Based Decoder} \label{sec:attention-decoder}

The decoder part is essential in the way that it translates the extracted semantic relationships into their corresponding captions. The decoder is an attention-based LSTM \cite{lu2017knowing}, which takes the word embedding vector ${\vec{w}^e_t}$, concatenated with the averaged semantic concepts vector $\vec{c}_{avg}=\frac{1}{N_g}\sum_{i=1}^{N_g}\vec{c}_i$, as input, which is denoted as $\vec{x}_t = [\vec{w}^e_t;\vec{c}_{avg}]$.

At each time step, when the decoder generates captions, the semantic relationship information that needs attention varies with different time steps. Inspired by the attention mechanism in \cite{lu2017knowing}, we design an attention LSTM sentence decoder by injecting all of the coarse-grained or fine-grained semantic relationship embedding features ($\vec{M}_k, k \in \{c,f\}$) into a LSTM with attention mechanism. Specifically, at each time step \textit{t}, according to the current hidden layer $\vec{h}_{t} \in \R^{{d}}$, the attention LSTM decoder firstly generates the attention distribution over the \textit{n} (\textit{n} is \textit{r} for $\vec{M}_c$ and 3\textit{r} for $\vec{M}_f$) parts of semantic relationship embedding features $\vec{M}_k, k \in \{c,f\}$:
\begin{align}
\vec{\alpha}_t = 
\text{softmax}(\vec{w}_{\alpha}\text{tanh}(\vec{W}_{M}{\vec{M}_k}^T \oplus \vec{W}_h\vec{h}_t)
\end{align}
where $\vec{W}_{M} \in \R^{d_a \times e}$, $\vec{W}_{h} \in \R^{d_a \times d}$ and $\vec{w}_{\alpha} \in \R^{d_a}$ are learnable parameters; $\oplus$ denotes the matrix-vector addition, which is calculated by adding the vector to each column of the matrix. $\vec{\alpha}_{t} \in \R^{n}$ is the attentive weight of $\vec{M}_{k}$ and the attentive relationship vector $\vec{r}_{t} \in \R^{e}$ can be defined as follows:
\begin{align}
\vec{r}_t = 
\vec{\alpha}_t\vec{M}_k
\end{align}
With a further combination with $\vec{h}_t$, the attentive relationship can predict the current output word:
\begin{align}
y_t \sim \vec{p}_t = \text{softmax}(\vec{W}_{p,r} \vec{r}_t + \vec{W}_{p,h}\vec{h}_t)
\end{align}
where $\vec{W}_{p,r} \in \R^{|D| \times e}, \vec{W}_{p,h} \in \R^{|D| \times d}$, and each value of $\vec{p}_t \in \R^{|D|}$ denotes the probability suggesting how likely the corresponding word in vocabulary $D$ is the current output word.

The approach mentioned above encourages the model to exploit all the available semantic relationship information. Thanks to the abundant and enriched information extracted by the \textit{Semantic Relationship Explorer}, the source semantic concept information turns into a deep image understanding.

\begin{table*}[t]

\centering
\footnotesize

\begin{tabular}{@{}l|c|c|c|c|c|c|c|c@{}}
\toprule
Model    & BLEU-1   & BLEU-2 & BLEU-3 & BLEU-4  & METEOR    & ROUGE-L  & CIDEr & SPICE     \\ \midrule [\heavyrulewidth]
 
Language Pivoting \cite{gu2018unpaired} & 46.2 & 24.0 & 11.2  & 5.4 & 13.2 & - & 17.7 & -  \\ 

Adversarial+Reconstruction \cite{Feng2018Unsupervised_ic} &  58.9  &40.3 & 27.0 & 18.6  & 17.9  & 43.1  & 54.9  & 11.1  \\

Graph-Align\ssymbol{2} \cite{gu2019unpaired}& 66.7 &48.0  & 31.5  & 21.1 &20.5  &46.2  &68.7  &14.5 \\ 

Graph-Align \cite{gu2019unpaired} &67.1 &47.8 &32.3 &21.5 &20.9 &47.2 &69.5 &15.0 \\ \midrule
[\heavyrulewidth]

Coarse-Grained SRE &  50.3 &  36.6 &  24.7 & 16.5  & 14.3 & 33.4 & 37.2  & 10.6 \\ 
Fine-Grained SRE & 57.5     &  38.9 & 27.1 & 19.7 & 17.4 &  41.9 & 49.7 &  13.3  \\ 
Graph-Align \cite{gu2019unpaired} + Fine-Grained SRE & \bf 67.8     & \bf 48.7 & \bf 33.6 &\bf 21.8  &\bf 22.1 & \bf 48.4 & \bf 75.7 &  \bf 16.1 \\
\bottomrule

\end{tabular}
\caption{Performance on the MSCOCO Karpathy test split \cite{karpathy2014deep} under the unpaired setting. All values are reported in percentage (\%). Higher is better in all columns. \ssymbol{2} denotes the results of the re-implemented Graph-Align for fair comparison. The proposed \textit{Fine-Grained SRE} has a better performance than the \textit{Coarse-Grained SRE}. Additionally, it is clear that the \textit{Fine-Grained SRE} approach also boosts the performance of Graph-Align, indicating that our approach learns very effective semantic relationship information even for scene-graphs.\label{tab:unpaired_setting}}
\end{table*}

\section{Implementation Details}

In this section, we will introduce how to train our model and how to use our model to generate image captions under the unpaired settings.

\subsection{Training Strategy}
\paragraph{Training in Language Domain}

Since there is no parallel data, we train the full model in the language domain, and use the method of reconstructing the original sentence to train the model, that is:

\textbf{Sentence} $\to$ \textbf{Semantic Concepts} $\to$ \textbf{Semantic Relationships} $\to$ \textbf{Semantic Relationship Embedding} $\to$ \textbf{Decoder} $\to$ \textbf{Sentence}

The \textit{Semantic Relationship Explorer} explores the Semantic Relationships $\vec{R}^{S}$ from the Semantic Concepts $\vec{C}^{S}$, which are extracted from the sentence $\vec{S}$, and obtain Semantic Relationship Embeddings $\vec{E}^{S}$ through a self-attention mechanism. Finally, the original sentence $\vec{S}$ is reconstructed by an \textit{Attention-based Sentence Decoder}.

\paragraph{Training Objectives}

Firstly, we train the full model by minimizing the Cross-Entropy (XE) Loss \cite{lu2018neural,chen2017scacnn,Yang2019NA,lu2017knowing,yao2017boosting,Yang2021NACF,gan2017semantic,wang2017skeleton}. Given a target ground truth sequence ${y}^{\text{*}}_{1:T}$ and a captioning model with parameters $\theta$, the goal is to minimize the following cross entropy loss:
\begin{align}
{L}_{CE}(\theta) = - 
\displaystyle{\sum_{t=1}^{T}}
\text{log}({p}_{\theta}({y}^{\text{*}}_{t} | {y}^{\text{*}}_{1:t-1}))
\end{align}
Recently, reinforcement learning methods have also been widely used for captioning model training \cite{anderson2018bottom,Yang2019SGAE,rennie2017self,yao2018exploring}. Consequently, we further employ a reinforcement learning (RL) loss to improve the performance of our model. In this case, the cross-entropy loss method is used to pre-train the model, after which our goal is to minimize the negative expected score as:
\begin{align}
{L}_{RL}(\theta) = - {E}_{{y}_{1:T} \sim {p}_{\theta}} [r({y}_{1:T})]
\end{align}
where $r$ is the score function  (e.g., CIDEr). Following the Self-Critical Sequence Training \cite{rennie2017self} (SCST), the gradient of ${L}_{RL}(\theta)$ can be approximated by
\begin{align}
{\nabla}_{\theta} {L}_{RL}(\theta) \approx - (r({y}^{s}_{1:T}) - r(\hat{y}_{1:T})){\nabla}_{\theta} log {p}_{\theta}({y}^{s}_{1:T})
\end{align}
where $r({y}^{s}_{1:T})$ is the score of a sampled caption ${y}^{s}_{1:T}$ and $r(\hat{y}_{1:T})$ suggests the baseline score of a caption which is generated by the current model using greedy decode. Through this gradient, sampled captions with higher CIDEr scores are more likely to be generated by the model because their corresponding probabilities are increased. Through this approach, we complete CIDEr optimization.

\subsection{Inference Strategy}

We use the full model to train the model in the language domain to generate image captions. The inference process is defined as follows:

\textbf{Image} $\to$ \textbf{Semantic Concepts} $\to$ \textbf{Semantic Relationships} $\to$ \textbf{Semantic Relationship Embedding} $\to$ \textbf{Decoder} $\to$ \textbf{Sentence}

Given an image $\vec{I}$, we can extract the Semantic Concepts $\vec{C}^{I}$ from $\vec{I}$ through the \textit{Semantic Concept Extractor}. Then, similar to sentence reconstruction process, we use the \textit{Semantic Relationship Explorer} to extract the Semantic Relationship $\vec{R}^{I}$, which embedded to \textit{Semantic Relationship Embedding}s $\vec{E}^{I}$. At last, $\vec{E}^{I}$ will use the decoder to generate captions.

\begin{table*}[t]

\centering
\footnotesize

\begin{tabular}{@{}l|c|c|c|c|c|c|c|c@{}}
\toprule
Model    & BLEU-1   & BLEU-2 & BLEU-3 & BLEU-4  & METEOR    & ROUGE-L  & CIDEr & SPICE     \\ \midrule [\heavyrulewidth]

\multicolumn{9}{@{}l}{\it ATT-FCN \cite{you2016image} }   \\ \midrule 
Baseline  &70.9 &53.7 & 40.2 & 30.4 & 24.3 &  53.3 & 95.6 & 18.6 \\  

\ + Proposal &  \bf 72.3   & \bf 55.6 & \bf 41.6  & \bf 32.0   & \bf 25.9   & \bf 54.8   & \bf 100.2   & \bf 19.2   \\ \midrule [\heavyrulewidth]

\multicolumn{9}{@{}l}{\it LSTM-A2 \cite{yao2017boosting} }   \\ \midrule 
Baseline  &73.3 &56.5 &42.7 &32.2 &25.3 &53.9 &99.1 &18.3\\

\ + Proposal &  \bf 77.2   & \bf 58.4  & \bf 44.6  & \bf 33.1   & \bf 25.9   & \bf 54.2  & \bf 103.5   & \bf 18.7   \\ \midrule [\heavyrulewidth]

\multicolumn{9}{@{}l}{\it LSTM-A3 \cite{yao2017boosting} }   \\ \midrule 
Baseline &73.5 &56.6& 42.9 &32.4& 25.5 &53.9 &99.8 &18.5   \\ 

\ + Proposal &  \bf 76.8   & \bf 57.9  & \bf 44.2  & \bf 33.2   & \bf 26.7   & \bf 54.5   & \bf 104.1   & \bf 19.1   \\ \midrule [\heavyrulewidth]

\multicolumn{9}{@{}l}{\it LSTM-A4 \cite{yao2017boosting} }   \\ \midrule 
Baseline   &72.1 &55.5 &41.7 &31.4 &24.9 &53.2 &95.7 &17.8 \\

\ + Proposal &  \bf 72.8   & \bf 56.4  & \bf 43.4  & \bf 32.1   & \bf 26.6   & \bf 54.6   & \bf 101.4   & \bf 18.4   \\ \midrule [\heavyrulewidth]

\multicolumn{9}{@{}l}{\it LSTM-A5 \cite{yao2017boosting} }   \\ \midrule 
Baseline  &73.4 &56.7 &43.0 &32.6 &25.4 &54.0 &100.2 &18.6\\

\ + Proposal &  \bf 75.2   & \bf 57.9  & \bf 44.2  & \bf 34.0   & \bf 26.7   & \bf 54.8   & \bf 108.0   & \bf 19.7   \\ \bottomrule

\end{tabular}
\caption{Evaluation of representative systems under the paired setting. The proposed approach can further improve the already strong baselines in all metrics, which demonstrates the generalization ability of our approach to a wide range of existing systems. The significant improvements come from the explored semantic relationships rather than the simple incorporation of the semantic concepts, which indicates its effectiveness in exploring semantic relationships. \label{tab:paired_setting}}
\end{table*}

\section{Experiments}

In this section, we will firstly describe a benchmark dataset for image captioning as well as some widely-used metrics and experimental settings. Then we will present the proposed model. \nocite{fenglin2019scaling}

\subsection{Datasets, Metrics and Settings}

Several datasets consist of images-caption pairs. We use the popular Microsoft COCO \cite{chen2015microsoft}  dataset to evaluate our reported results. This dataset contains 123,287 images and each image is paired with 5 sentences. To make a fair comparison, we use the widely-used splits in the work of Karpathy and Li \cite{karpathy2014deep} to report our results. There are 5,000 images each in validation set and test set, and 566,435 sentences for training full model.

We test the model performance with MSCOCO captioning evaluation toolkit \cite{chen2015microsoft}. It reports the widely-used automatic evaluation metrics SPICE, CIDEr, BLEU, METEOR and ROUGE. SPICE \cite{anderson2016spice} is based on scene graph matching and CIDEr \cite{vedantam2015cider} is based on n-gram matching. These two metrics are specifically designed to evaluate image captioning systems. They both incorporate the consensus of a set of references for an example. BLEU \cite{papineni2002bleu} and METEOR  \cite{lin2003automatic,banerjee2005meteor}  are originally designed for machine translation evaluation, while ROUGE \cite{lin2004rouge} is proposed for automatic evaluation of the extracted text summarization. A related research suggests that SPICE correlates the best with human judgments and does especially well in judging detailedness, where the other metrics present negative correlations; CIDEr and METEOR follows with no conspicuous superiority, then comes ROUGE-L, and BLEU-4, in that order \cite{anderson2016spice,vedantam2015cider}. Among them, SPICE and CIDEr are specifically designed to evaluate image captioning systems and will be the main considered metrics.

Following convention, we replace caption words appearing less than 5 times in the training set with the commonly unknown word token UNK. This has resulted in 9,487 words for MSCOCO. The size of the word embedding $e$ is 512 and the hidden size $d$ of the LSTM is 512. $N$, the number of heads in multi-head attention is set to 8 and ${d}_{ff}$, the feed-forward network dimension is set to 2048. Besides, the hidden layer size ${d}_a$ is 350. The relationship embedding has 30 rows (the $r$), which is similar to \cite{lin2017Embedding}. 
Following \cite{gu2019unpaired}, we train the \textit{Semantic Concept Extractor} on the Visual Genome \cite{krishna2017visualgenome} dataset.
Only top 20 semantic concepts are selected for each
image.  We also share the semantic concept embedding and the input word embedding in implementation. During the training process, firstly, we train the model with the cross-entropy loss for 15 epochs. Then, in reinforcement learning, we train the entire model with the batch size of 50 and use Adam for parameter optimization. For full model, the learning rate is 4e-4. The $\beta_1, \beta_2$ are respectively set to 0.8 and 0.999. During sampling for MSCOCO dataset, we apply beam-search with a beam size of 5.

\subsection{Results}

\subsubsection{Under the Unpaired Setting}
In this section, in order to show the advantages of our method, firstly we compare our approach with the current unpaired image captioning methods, including the recently proposed Language Pivoting \cite{gu2018unpaired}, Adversarial+Reconstruction \cite{Feng2018Unsupervised_ic} and Graph-Align \cite{gu2019unpaired}, which are the most advanced on the MSCOCO dataset in comparable settings. As shown in Table~\ref{tab:unpaired_setting}, in terms of SPICE, our proposed \textit{Fine-Grained SRE} is competitive with all baselines, which use complex models and schemes. 
This is reasonable because both semantic concepts and semantic relationships are high-level and understandings of image \cite{wu2016what}. 
Additionally, we boost the performance of Graph-Align, which indicates vision domain and language domain can be connected by our approach effectively.
As expected, the \textit{Fine-Grained SRE} has a better effect than the \textit{Coarse-Grained SRE}. This indicates that the \textit{Fine-Grained SRE} helps improve the quality of the extracted semantic relationships by extracting relevant relationships in more precise ranges.

\subsubsection{Under the Paired Setting}
We have done experiments to further illustrate the advantages of our method. Under the paired setting, we use five strong baselines without exploring the semantic relationships for semantic concepts. We replace the source semantic concept information of the baselines with extracted semantic relationship information, which is extracted by \textit{Fine-Grained Semantic Relationship Explorer}. Since our focus is to provide semantic relationship information, we preserve the original settings for all baselines.

ATT-FCN \cite{you2016image} attends semantic concepts in all decoding steps. LSTM-A \cite{yao2017boosting} uses a series of models (LSTM-A2,3,4,5) to incorporate the semantic concepts. We will study the effect of our method on them. LSTM-A3 takes semantic concepts as input at the first decoding step and visual features at the second step. LSTM-A4 presents textual concepts to the decoder, leaving visual features for the following steps; Contrary to LSTM-A3 and LSTM-A4, LSTM-A2 and LSTM-A5 reverse this order by firstly presenting visible features, respectively. 

As shown in Table~\ref{tab:paired_setting}, \textit{Fine-Grained SRE} successfully promotes all baseline systems. It has brought improvements up to 8\% and 6\% in terms of CIDEr and SPICE, respectively, which further demonstrates its wide-range generalizing ability and indicates its effectiveness in exploring semantic relationships, which is less prone to the variations of model structures, hyper-parameters (e.g., learning rate and batch-size), and learning paradigm.

To conclude, we not only performs better than existing methods under the unpaired setting, but manage to greatly improve the baselines under the paired setting.

\begin{table*}[t]

\centering
\footnotesize

\begin{tabular}{@{}l|c|c|c|c|c|c|c|c@{}}
\toprule
Model    & BLEU-1   & BLEU-2 & BLEU-3 & BLEU-4  & METEOR    & ROUGE-L  & CIDEr & SPICE     \\ \midrule [\heavyrulewidth]
 
GT-ATT \cite{wu2016what} & 82.3 & 71.7 & 54.2  & 44.9 & 30.7 &  59.7 & 143.9 & 24.1  \\  \midrule [\heavyrulewidth]

GT-SRE & 84.7 & 72.9 & 57.8  & 46.4 & 32.0 &  61.5 & 154.1 & 26.2   \\ 

GT-SRE-Att & 90.6     & 80.4 & 65.9 & 57.2  & 35.5 & 66.9  & 161.5 &  27.4 \\ 
GT-SRE-Emb-Att (Full Model) & \bf 92.4 & \bf 86.9 & \bf 74.4 & \bf 63.1 & \bf 36.7 & \bf 69.1 & \bf 174.4 & \bf 28.3 \\ 
\bottomrule

\end{tabular}
\caption{Performance of the proposed models using the ground-truth semantic concepts on MSCOCO. The attention mechanism (\textit{Att}) and the embedding mechanism (\textit{Emb}) can promote the baseline in all metrics. Additionally, we report the performance of a representative system ATT \cite{wu2016what}. As we can see, our approach outperforms the ATT substantially in all metrics, which further demonstrates the effectiveness of our approach. \label{tab:gt-analysis}}
\end{table*}

\begin{table*}[t]

\centering

\footnotesize
\begin{tabular}{@{}l c c c c c c c c c c c@{}}
\toprule
\multirow{2}{*}{Methods}   & \multirow{2}{*}{BLEU-4}     & \multirow{2}{*}{METEOR}     & \multirow{2}{*}{ROUGE-L}    & \multirow{2}{*}{CIDEr}   &  \multicolumn{7}{c}{SPICE}     \\  \cmidrule(lr){6-12} 
 & & && & All &Objects &Attributes &Relations &Color &Count &Size 
\\ \midrule

Graph-Align   & 21.1 &20.5  &46.2  &68.7  &14.5 & 28.8 &  3.8 & 3.8  & 1.7  &  0.7 &  1.9\\

Graph-Align w/ ``Attribute''  & \bf 21.5 & 20.9 & \bf 47.0 &\bf 71.2 & 15.0 & 29.2 &  \bf  5.6 & 3.4  &\bf  3.1  & \bf  2.6 & \bf  2.3 \\

Graph-Align w/ ``Object''  &   21.3 & \bf  21.4 & 46.7 &  70.1 & \bf  15.6 &  \bf 30.7 &  4.2 & 4.3  & 2.4  &  1.5 &  2.0 \\

Graph-Align w/ ``Relation'' & 20.7 & 20.4 & 46.4 & 69.8 & 14.7 & 29.0 &  3.6 & \bf  4.5  & 2.0  & 1.1 &  1.9 \\ \midrule  [\heavyrulewidth]

Full Model &  21.8 &  22.1 &   48.4 & 75.7 &  16.1  &   31.0     &  5.8     &  4.9     & 3.4     &  3.7     &  2.5   \\
\bottomrule
\end{tabular}
\caption{Results of incremental analysis of our proposed approach under the unpaired setting. For a better understanding of the differences, we further list the breakdown of SPICE F-scores. \textit{``w/ Attribute''}, \textit{``w/ Object''} and \textit{``w/ Relation''} stand for the baseline further equipped with the \textit{``Attribute'' Aggregator}, \textit{``Object'' Aggregator} and \textit{``Relation'' Aggregator}, respectively. The bold numbers are best numbers before applying full model. We can see that the \textit{``w/ Attribute''} has a higher attributes, colors, count and size scores than the other baselines. The \textit{``w/ Object''} promotes the baseline in objects scores and the \textit{``w/ Relation''} reaches better scores in relations. As we can see, incorporating all the sub-modules (i.e., \textit{{Full Model}}) directly on the baseline leads to overall improvements.
\label{tab:ablation}}
\end{table*}
\begin{table*}[]
\footnotesize
\centering
\begin{tabular}{|@{}p{.12\linewidth}|p{.193\linewidth}|p{.193\linewidth}|p{.193\linewidth}|p{.193\linewidth}@{}|}
\hline

\includegraphics[width=1\linewidth,height=1.5\linewidth]{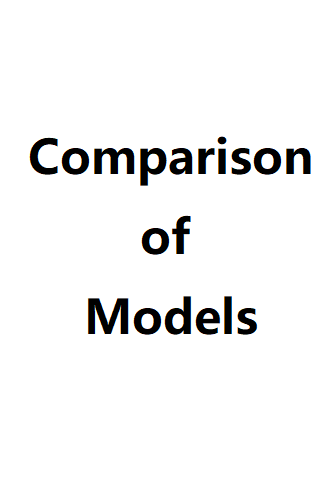} &\includegraphics[width=1\linewidth,height=1\linewidth]{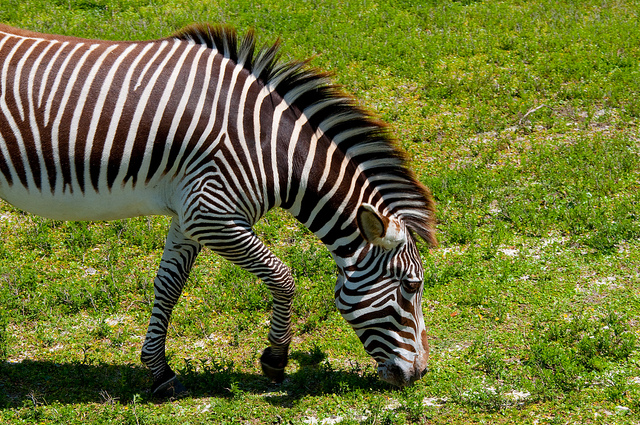} &\includegraphics[width=1\linewidth,height=1\linewidth]{284910.jpg} &\includegraphics[width=1\linewidth,height=1\linewidth]{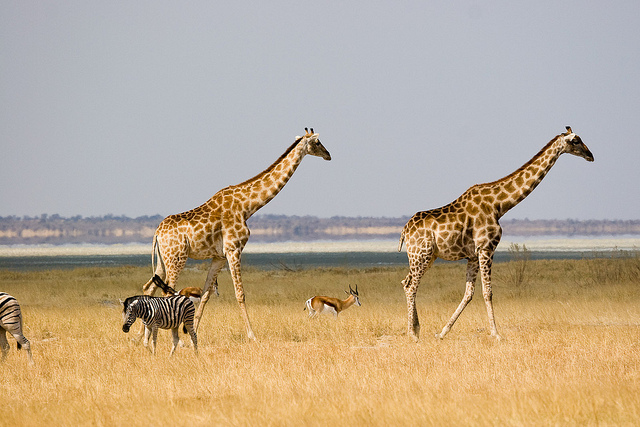}   &\includegraphics[width=1\linewidth,height=1\linewidth]{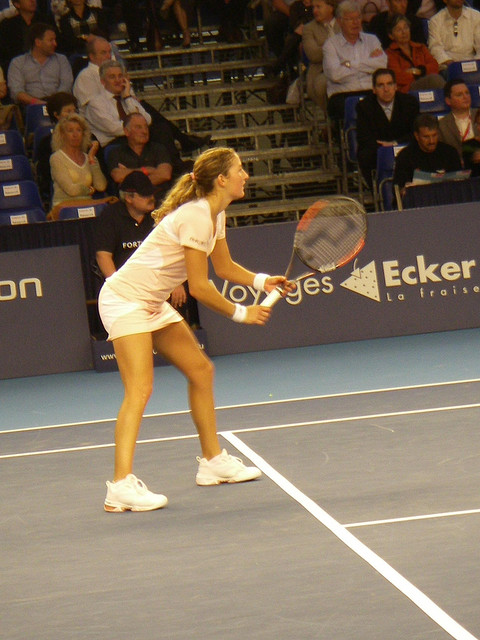}   \\ \hline

\textbf{\ Reference}    &\begin{tabular}[c]{@{}l@{}} a zebra grazing on grass in a \\ green open field. \end{tabular}                          &\begin{tabular}[c]{@{}l@{}}  a woman is outside playing \\ with her dogs in the grass. \end{tabular}  &\begin{tabular}[c]{@{}l@{}} a herd of wild animals walking \\ across a large dry field. \end{tabular}  &\begin{tabular}[c]{@{}l@{}} a woman playing tennis on a \\ tennis court while people \\ watch.  \end{tabular}      \\ \hline

\textbf{\ Graph-Align}  &\begin{tabular}[c]{@{}l@{}} an animal standing on a field.  \end{tabular}  &\begin{tabular}[c]{@{}l@{}} a woman standing on a field. \end{tabular} &\begin{tabular}[c]{@{}l@{}} an animal standing on a field.  \end{tabular}    &\begin{tabular}[c]{@{}l@{}}  a woman standing on a field. \end{tabular}      \\ \hline

\textbf{\ \  w/ ``Attribute''} &\begin{tabular}[c]{@{}l@{}} an animal standing on lush \\ green grass field. \end{tabular}          &\begin{tabular}[c]{@{}l@{}} a beautiful woman standing on \\ a grass field. \end{tabular} &\begin{tabular}[c]{@{}l@{}}a herd of animals standing on \\ a dry field. \end{tabular}    &\begin{tabular}[c]{@{}l@{}} a woman standing on a tennis \\ court. \end{tabular}    \\ \hline

\textbf{\ \ w/ ``Object''} &\begin{tabular}[c]{@{}l@{}} a zebra standing on a field \\ with grass. \end{tabular}          &\begin{tabular}[c]{@{}l@{}}  a woman in shirt and dogs \\ playing on a grass field. \end{tabular} &\begin{tabular}[c]{@{}l@{}} zebras and giraffes standing on \\ a dry field. \end{tabular}    &\begin{tabular}[c]{@{}l@{}} a woman with a racket \\ standing on a tennis court.  \end{tabular}    \\ \hline

\textbf{\ \ w/ ``Relation''} &\begin{tabular}[c]{@{}l@{}} an animal standing on a field \\ eating grass. \end{tabular}          &\begin{tabular}[c]{@{}l@{}} a woman wearing shirt standing \\ on a grass field with dogs \\ nearby. \end{tabular} &\begin{tabular}[c]{@{}l@{}} animals walking in a field. \end{tabular}    &\begin{tabular}[c]{@{}l@{}} a woman holding a racket \\ standing on a tennis court. \end{tabular}    \\ \hline

\textbf{\ Full Model} &\begin{tabular}[c]{@{}l@{}} a zebra eating lush green grass \\ in a grass field. \end{tabular}                       &\begin{tabular}[c]{@{}l@{}} a woman wearing black shirt is \\ playing with dogs on a green \\ grass field. \end{tabular} &\begin{tabular}[c]{@{}l@{}} giraffes and zebras standing on  \\ a dry grass field. \end{tabular}    &\begin{tabular}[c]{@{}l@{}} a woman wearing white shirt \\ holding a tennis racket on the \\ court. \end{tabular}   \\ \hline
\end{tabular}
\caption{Examples of the captions generated by different methods. The first line is the input image. The second and the third lines are the ground truth and the captions generated by the Graph-Align. The fourth, the fifth and the sixth lines are captions generated by adding \textit{``Attribute'' Aggregator}, \textit{``Object'' Aggregator} and \textit{``Relation'' Aggregator}, respectively. From the captions, we can find that the \textit{``Attribute'' Aggregator} helps the baseline to generate more detailed captions in attributes and colors for each object. The \textit{``Object'' Aggregator} results in more comprehensiveness in objects, and the \textit{``Relation'' Aggregator} helps the baseline to establish more reasonable relationships among objects. The \textit{Full Model}, which incorporates all sub-modules, is able to generate more complete captions that are detailed in the objects, attributes, relations and colors.}
\label{fig:vis}
\end{table*}
\section{Analysis}
In this section, we will first analyze the contribution of each component in the proposed method. Then examples and error analysis under the unpaired setting are given to demonstrate the strengths and areas for improvement of our model. The following analysis are conducted on the the proposed \textit{Fine-Grained SRE}.

\subsection{Quantitative Analysis}

The generic captioning performance can be affected by the \textit{\textit{Semantic Relationship Embedding}} and \textit{Attention-based Sentence Decoder} we adopted. Therefore, we first assume that ground truth semantic concepts are fixed and then analyze them. The results are shown in Table~\ref{tab:gt-analysis}. In addition, because of our main contribution is proposing the \textit{Semantic Relationship Explorer}, we  select the Graph-Align to conduct a series of experiments under real unpaired situations to evaluate the contribution of sub-modules. The results are shown in Table~\ref{tab:ablation}. Meanwhile, we have also listed results of SPICE sub-categories to make it easier to analyze the quality and differences of the captions.

\paragraph{Contribution of \textit{Semantic Relationship Embedding} and Attention-based Sentence Decoder}

As shown in Table~\ref{tab:gt-analysis}, the comparison of the three models indicates the upper limit of the performance of attention mechanism (\textit{Att}) and embedding mechanism (\textit{Emb}). It is demonstrated that the \textit{Attention-based Sentence Decoder} (\textit{GT-SRE-Att}) outperforms the non-attention model (\textit{GT-SRE}) in all metrics. This proves that weighing over features can improve the global dependency of features. The performance of model is greatly improved when it is combined with \textit{Semantic Relationship Embedding} (\textit{GT-SRE-Emb-Att}). This indicates that \textit{Semantic Relationship Embedding} can further extract multiple features from semantic relationships, which strengthens its representing ability.

Furthermore, our full model performs better than \textit{GT-ATT}. The latter directly uses semantic concepts at all steps instead of exploring the semantic relationships for semantic concepts. This proves the necessity to build relationships between semantic concepts.

\paragraph{Effect of ``Attribute'' Aggregator.}

As shown in Table~\ref{tab:ablation}, if we incorporate our model with the \textit{``Attribute'' Aggregator} and aggregate correlated attributes in each specific object, it will have strengthened attribute relationship representations. Consequently, the object united with the strengths of attribute collocations can produce a balanced improvement. This is especially true in \textit{attributes}, \textit{color} and \textit{count} score. It can also promote the decoder to generate accurate captions.

\paragraph{Effect of ``Object'' Aggregator.}

Just as we have expected, \textit{``Object'' Aggregator} performs well in connecting related objects in the semantic concepts according to a relation word, which is proved by the increased scores in \textit{objects}. The object collocations use the ``relation'' word as a pivot to extract constrained collocations, which distills more detailed relationship information between objects, promoting the decoder to generate comprehensive captions.

\paragraph{Effect of ``Relation'' Aggregator.}

As is shown in Table~\ref{tab:ablation}, the base model, which directly incorporates \textit{``Relation'' Aggregator}, have improved \textit{relations}. This indicates that the introduction of relation collocations has induced more powerful relation information to the decoder. The decoder therefore establishes more fine-grained relationships around a specific object.
 
\paragraph{Full Model.}

As we expected, each component, with different functions, brings about improvements from different aspects to the model. Consequently, their advantages are united to produce abundant and enriched information. This achieves deep image understanding, brings an overall improvement (see Table~\ref{tab:ablation}) and results in generating comprehensive and detailed captions (see Figure~\ref{fig:vis}).

\subsection{Qualitative Analysis}

In the Figure~\ref{fig:vis}, we list examples to intuitively show the differences between models. As is shown, all models can generate fluent and descriptive sentences of input images. However, different models present different amount of semantic relationship information. The \textit{w/ Object} has more objects but lacks details, e.g. color and number. The \textit{w/ Attribute} generates more detailed attributes and color. The \textit{w/ Relation} performs well in portraying the relations but worse in attributes. The proposed \textit{Full Model}, compared with the models mentioned above, helps the baseline to maintain good balances as well as generates more complete and coherent captions.

The reason for this phenomenon is \textit{``Object'' Aggregator} can automatically extend its focus in a specific relation and find related objects to establish a relationship group. \textit{``Attributes'' Aggregator} learns attribute-word collocations from various visual attributes and rich semantic information. \textit{``Relation'' Aggregator} learns to automatically extend its focus in a specific object and look for related relations. Based on rich semantic relationship information, the decoder can retrieve information through attention mechanism. It will also decide on the most probable next word. 

The qualitative analysis indicates that the semantic relationship information, extracted by \textit{Semantic Relationship Explorer}, is much more aggregated than the source semantic concept information, ensuring the focus of current caption generation.

\begin{figure}[t]

\centering
\includegraphics[width=0.85\linewidth]{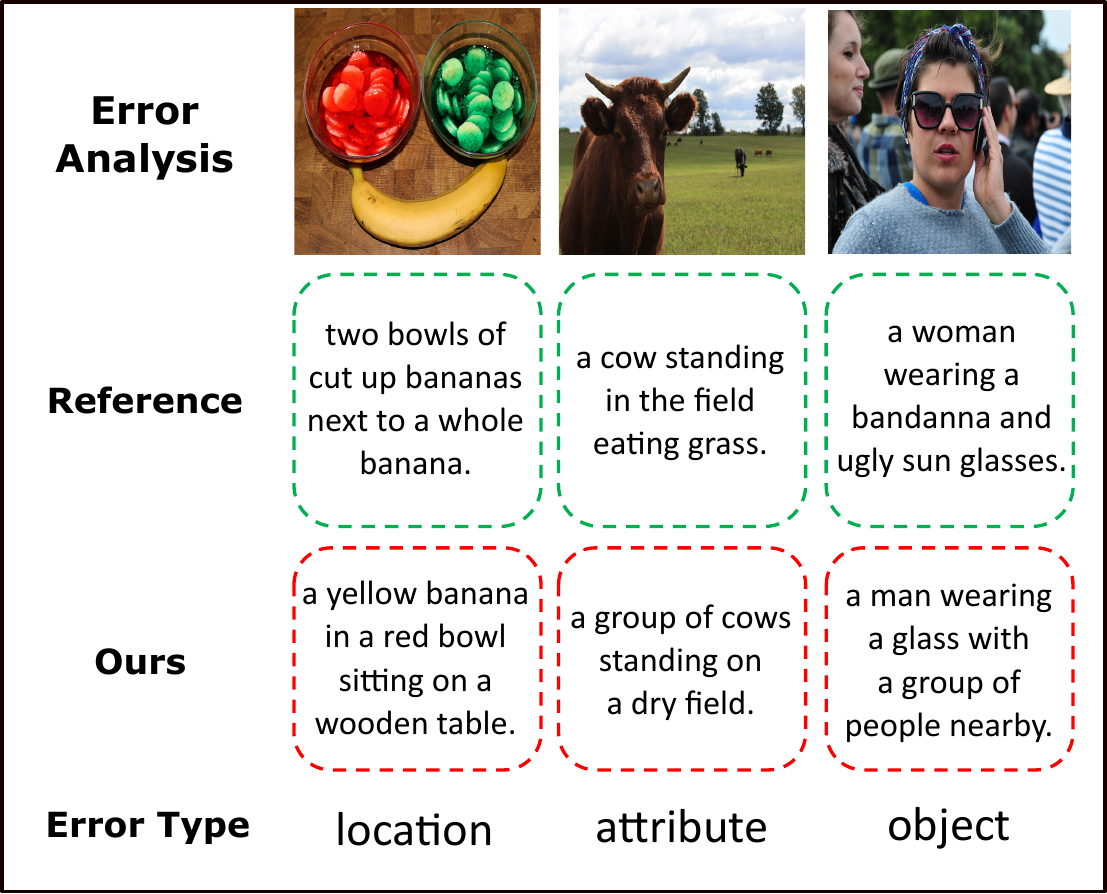}

\caption{Illustrations of the error analysis of the proposal.}
\label{fig:error_analysis}
\end{figure}

\subsection{Error Analysis}

In this section, we conduct error analysis using the proposed (full) model on the test set to provide insights on how the model may be improved. We find some generated captions that do not meet the standard. There are mainly three types of errors, i.e., location (45.9\%), attribute (21.7\%), and object (18.4\%), with the left (14.0\%) falling into other categories. In the Figure~\ref{fig:error_analysis}, we also give some examples.

Location error means that the model describes an irrelevant location relationship for an object (e.g., mislocating \textit{banana}). 
Attribute error occurs when there are a lot of attributes, which can respectively establish logical relationships with a specific object. Therefore, the model mistakenly treats an incorrect relationship as grounded. In the given example, \textit{field} can establish a relationship with either \textit{dry} or \textit{grass}, but only \textit{grass field} is correct as shown in the picture. 
Object error occurs when there are incorrect extracted semantic concepts. The model treats the semantic concept as grounded in the image. In the given example, the incorrect semantic concept is \textit{man}. 

The challenging part is that these semantic relationships, which are extracted by the \textit{Semantic Relationship Explorer}, are logically correct, otherwise the proposed model will choose other semantic relationships. A more powerful and accurate \textit{Semantic Concept Extractor} may be needed to solve the problem, but it's unlikely to be completely avoided.

\section{Conclusions}

In this work, we have proposed a framework for unpaired image captioning and a \textit{Semantic Relationship Explorer} (SRE). Without any parallel data, SRE extracts abundant and rich semantic relationship information for image captioning. Our method employs the semantic relationships to bridge the gap between vision and language domains. It is a powerful basis for the image description. We also use \textit{\textit{Semantic Relationship Embedding}} to get more powerful relationship features. Meanwhile the \textit{Attention-based Sentence Decoder} is adopted to improve the global dependency of features. Our proposal is validated by experiments on the MSCOCO image captioning dataset. Remarkably, the proposed method outperforms the existing methods under the unpaired setting. In addition, it manages to boost five strong baselines to achieve a significant improvement under the paired setting. As shown in the quantitative and qualitative analysis, generated captions achieve a great balance among ``object'', ``attribute'' and ``relation'', which are more complete and coherent compared to other existing models.

\bibliographystyle{IEEEtran}
\bibliography{IEEEexample}

\end{document}